\documentclass[letterpaper, 10 pt, journal, twoside]{IEEEtran}
\ifCLASSINFOpdf
\else
\fi

\usepackage{graphics} 
\usepackage{epsfig} 
\usepackage{mathptmx} 
\usepackage{times} 
\usepackage{amsmath} 
\usepackage{amssymb}  
\usepackage{graphicx}
\usepackage{caption, subcaption}
\usepackage{arydshln}

\usepackage{booktabs}
\graphicspath{ {./images/} }
\usepackage[dvipsnames]{xcolor}
\definecolor{myYellow}{RGB}{248,231,28}
\definecolor{myRed}{RGB}{255,124,140}
\definecolor{colorSky}{RGB}{39, 159, 216}
\definecolor{colorPlants}{RGB}{7, 130, 42}
\definecolor{colorGround}{RGB}{82, 91, 96}
\newcommand\crule[3][black]{\textcolor{#1}{\rule{#2}{#3}}}

\usepackage{float}
\usepackage{xspace}
\usepackage[dvipsnames]{xcolor}

\newcommand{\ie}{\emph{i.e.}\xspace}
\newcommand{\eg}{\emph{e.g.}\xspace}

\newcommand{\Fig}{Fig.\xspace}

\newcommand{\etal}{\emph{et al.}\xspace}
\newcommand{\revised}{\textcolor{black}}{}
\begin{document}
%
\title{Unsupervised Image Segmentation by \\ Mutual Information Maximization and \\ Adversarial Regularization}
%
%
%

\author{S. Ehsan Mirsadeghi$^{1}$, Ali Royat$^{2}$,  Hamid Rezatofighi$^{3}$

\thanks{Manuscript received: February, 25, 2021; Revised April, 22, 2021; Accepted June, 14, 2021.}
\thanks{This paper was recommended for publication by Editor FirstName A. EditorName upon evaluation of the Associate Editor and Reviewers' comments.} 
\thanks{$^{1}$S. Ehsan Mirsadeghi is with the Faculty of Electrical Engineering, AmirKabir University of Technology, Iran
        {\tt\small e.mirsadeghi@aut.ac.ir}}%
\thanks{$^{2}$Ali Royat is with the Faculty of Electrical Engineering, Sharif University of Technology, Iran
        {\tt\small royat.ali@ee.sharif.edu}}%
\thanks{$^{3}$Hamid Rezatofighi is a lecturer at the Department of Data Science and AI, Faculty of Information Technology, Monash University, Australia.
        {\tt\small Hamid.Rezatofighi@monash.edu}}
\thanks{Digital Object Identifier (DOI): see top of this page.}
}
%
%

\markboth{IEEE Robotics and Automation Letters. Preprint Version. Accepted June, 2021}
{Mirsadeghi \MakeLowercase{\textit{et al.}}: Unsup. Image Seg. by Mutual Information Maximization and Adv. Regularization} 

%



\maketitle

\begin{abstract}
Semantic segmentation is one of the basic, yet essential scene understanding tasks for an autonomous agent. The recent developments in supervised machine learning and neural networks have enjoyed great success in enhancing the performance of the state-of-the-art techniques for this task. However, their superior performance is highly reliant on the availability of a large-scale annotated dataset. 
In this paper, we  propose  a  novel fully  unsupervised  semantic  segmentation  method, the so-called \textbf{In}formation \textbf{M}aximization and \textbf{A}dversarial \textbf{R}egularization \textbf{S}egmentation  (InMARS). Inspired by human perception which parses a scene into perceptual groups, rather than analyzing each pixel individually, our proposed approach first partitions an input image into meaningful regions (also known as superpixels). Next, it utilizes Mutual-Information-Maximization followed by an adversarial training strategy to cluster these regions into semantically meaningful classes.
To customize an adversarial training scheme for the problem, we incorporate adversarial pixel noise along with spatial perturbations to impose photometrical and geometrical invariance on the deep neural network. Our experiments demonstrate that our method achieves the state-of-the-art performance on two commonly used unsupervised semantic segmentation datasets, COCO-Stuff, and Potsdam.
\end{abstract}


%
\IEEEpeerreviewmaketitle

\section{Introduction}
%
%
%
%
\IEEEPARstart{M}{any} autonomous agents, \eg mobile robots and driverless vehicles, rely on scene understanding tasks for navigation and interaction with the environment and semantic segmentation is one of the basic, yet most important visual perception tasks~\cite{DeepLab, RefineNet, HRNet,  PSPNet}. In recent years, the significant developments in machine learning and neural networks have been conducive to supervised semantic segmentation regime~\cite{DeepLab, RefineNet, FCN}. However, the superior performance of these supervised techniques is highly subject to the availability of a large-scale and high-quality annotated dataset, and providing such a pixel-level annotation for a dataset from a novel domain, \eg satellite or underwater images, can be a very challenging and laborious task~\cite{ARL5, ARL17, IIC38}.

Recently unsupervised learning techniques such as deep clustering methods show promises in dealing with image-level tasks such as unsupervised image classification problems~\cite{DEC, VaDE, InfoGAN, IMSAT, AE+CE}. More specifically, maximizing mutual information plays a remarkable role in the core of recent works, and many algorithms have successfully utilized this concept for unsupervised image classification~\cite{DEC, VaDE, InfoGAN, IMSAT, JULE, DAC}. However, these techniques cannot be naively applied to semantic segmentation tasks, which require spatially smooth and consistent label predictions at the pixel-levels.

\begin{figure}
    \centering
    \begin{subfigure}[t]{0.11\textwidth}
      \includegraphics[width=\textwidth]{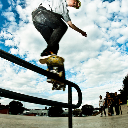}
    \end{subfigure}
    \begin{subfigure}[t]{0.11\textwidth}
      \includegraphics[width=\textwidth]{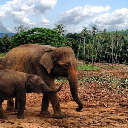}
    \end{subfigure}
    \begin{subfigure}[t]{0.11\textwidth}
      \includegraphics[width=\textwidth]{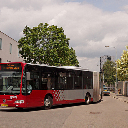}
    \end{subfigure}
    \begin{subfigure}[t]{0.11\textwidth}
      \includegraphics[width=\textwidth]{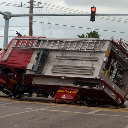}
    \end{subfigure}

    \par\smallskip
    \centering
    \begin{subfigure}[t]{0.11\textwidth}
      \includegraphics[width=\textwidth]{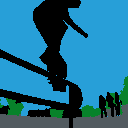}
    \end{subfigure}
    \begin{subfigure}[t]{0.11\textwidth}
      \includegraphics[width=\textwidth]{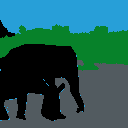}
    \end{subfigure}
    \begin{subfigure}[t]{0.11\textwidth}
      \includegraphics[width=\textwidth]{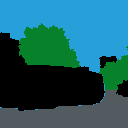}
    \end{subfigure}
    \begin{subfigure}[t]{0.11\textwidth}
      \includegraphics[width=\textwidth]{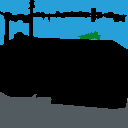}
    \end{subfigure}
    
    \par\smallskip
    \centering
    \begin{subfigure}[t]{0.11\textwidth}
      \includegraphics[width=\textwidth]{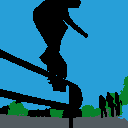}
    \end{subfigure}
    \begin{subfigure}[t]{0.11\textwidth}
      \includegraphics[width=\textwidth]{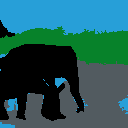}
    \end{subfigure}
    \begin{subfigure}[t]{0.11\textwidth}
      \includegraphics[width=\textwidth]{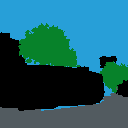}
    \end{subfigure}
    \begin{subfigure}[t]{0.11\textwidth}
      \includegraphics[width=\textwidth]{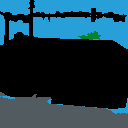}
    \end{subfigure}
    
    \caption{\textbf{Segmentation Result}. (first row) Some examples from COCO-Stuff datasets~\cite{Cocostuff}, (second row) their truth, and (third row) the predicted segmentation masks using our proposed method (InMARS$^{+}$).}
    \label{fig:teaser:coco-st3}
  \end{figure}
  
Recently, there have been few attempts to tackle this challenging problem by conducting various strategies and domain knowledge~\cite{WNet, IIC, SegSort, ARL}. However, these techniques are more susceptible to degenerate/trivial solutions, and their superior performance are highly dependent on their incorporated data augmentation strategy. \par
This paper introduces a novel pixel-wise and fully unsupervised learning method for semantic segmentation. Our proposed approach uses an adversarially trained information maximization technique over the features extracted from a set of segmented regions, also known as superpixels.
By extracting features from these grouped pixels, we, indeed, reformulate unsupervised image segmentation as a sub-region unsupervised classification task and transform decision space from pixel-level into sub-region domain. Next, we utilize Mutual Information Maximization for maintaining maximum information between input and output, which enforces semantically similar and different data to be assigned to the same and distinct clusters, respectively. We also incorporate an adversarial training strategy to act as a regularizer for our model, helping it learn a robust and generalizable representation for clustering the extracted features. To customize our learning regime for the segmentation task, we also propose a novel adversarial strategy by bundling a learnable adversarial pixel intensity noise with a set of random geometric perturbations on the superpixels.
We test and evaluate the proposed method on four subsets of two commonly used unsupervised image segmentation benchmark datasets: Potsdam \cite{Potsdam}, and COCO-stuff \cite{Cocostuff}. We demonstrate the superior performance of our proposed method, outperforming the existing state-of-the-art methods on these datasets. Moreover, we perform an ablation study to show each component's contribution to the proposed method. In summary, the main contributions of this work can be summarized as follows:

\begin{enumerate}
\itemsep0em
\item We propose a new unsupervised image segmentation approach based on adversarial training and mutual information,
\item We introduce a novel sub-region-level training strategy for image segmentation and incorporate customized adversarial pixel noise followed by spatial perturbations to make the model robust to the  photometrical and geometrical variations,
\item Our approach achieves the state-of-the-art results on commonly used unsupervised semantic segmentation datasets.
\end{enumerate}\par

%

\section{Related works} \label{sec:ReWo}
\textbf{Supervised Semantic Segmentation}: The recent advances in deep learning and convolutional neural architectures, along with the availability of large-scale annotated datasets, have been helpful to the impressive performance of the recent semantic segmentation algorithms~\cite{ReviewSS2020}. Fully Convolutional Network (FCNs)~\cite{FCN} is a successful generation of convolutional neural network (CNN) architectures that achieved a significant improvement over classical approaches. Several notable architectural variants of FCNs are later proposed by exploiting various strategies. For instance, \cite{RefineNet} and \cite{PSPNet} proposed different multi-scale and pyramid FCN architectures to learn multi-scale feature representation resulting in a better segmentation performance for various region sizes' resolution. Chen \etal~\cite{DeepLab} also proposed atrous spatial pyramid pooling to segment objects at multiple scales robustly while applying Conditional Random Field (CRF) additionally at the end of the network to smooth out the pixel-wise label predictions. Jun \etal~\cite{DANet} suggested an attention mechanism to integrate local features with their global dependencies. Recently, \cite{HRNet} introduced a computational tractable multi-stage architecture to extract a fast but high-resolution feature representation. Despite the progressive success of supervised techniques, their great performance depends on the availability of large-scale annotated data.\par

\textbf{Deep Clustering}: 
Data clustering is a classical pattern recognition problem, and several model-based approaches, \eg K-means~\cite{He_2013_CVPR}, Gaussian Mixture~\cite{GMM-Cl},  Kernel-based~\cite{NIPS2004_2602} and spectral clustering~\cite{On-spectral-clustering}, have been developed to tackle this fundamental problem. However, their performance considerably degrades on highly non-linear data and/or in a high dimensional space, \eg raw and unprocessed images. Deep learning has provided an opportunity for these clustering techniques to overcome their drawbacks by clustering on a proper embed (and possibly low-dimensional) feature space, encoded by these neural models. Deep Embedded Clustering (DEC)~\cite{DEC} is one of the pioneering works, which combines one of the classical clustering methods with deep learning. This method combines a reconstruction loss on an Auto-Encoder (AE) with a clustering objective on the encoded features to construct and cluster the rich low-dimensional encoded features, simultaneously. Other similar approaches also use AE reconstruction loss but followed by a K-means~\cite{kmean+AE}, or a cross-entropy objective between multiple network predictions~\cite{AE+CE} to perform the clustering task on the deep embed space. The other generative models such as Variational AE (VAE) ~\cite{VaDE} and InfoGAN~\cite{InfoGAN} are also used in combination with clustering objectives to learn a clustering-friendly latent space. 

Unlike the earlier works, IMSAT~\cite{IMSAT} considers the clustering task as a discrete representation problem, where each discrete representation would be a code for the assignment of input data to a specific cluster. Then an objective is defined to maximize mutual information between inputs and outputs while regularizing this representation by an adversarial perturbation. In spite of their success in the data clustering problems, these deep clustering methods cannot be naively used for the unsupervised semantic segmentation problem where the task is to cluster semantically same pixels with high spatial/local dependencies using fine-grained deep features.\par

\textbf{Unsupervised Semantic Segmentation}:
Being a very challenging problem, there have been a few recent attempts to tackle unsupervised image segmentation. 
Hwang \etal~\cite{SegSort} proposed a method by combining two sequential clustering modules for both pixel-level and segment-level to perform the task. Notwithstanding its satisfactory performance, this approach relies on mean-shift clustering, making it susceptible to the degeneracy issue, \ie assigning random labels to different candidates. 
Generative model-based approaches such as~\cite{ARL4, ReDrawGAN} generate a semantic mask for an image by perturbing~\cite{ARL4} or redrawing~\cite{ReDrawGAN} the generated foreground and background masks. Despite their great performance, their application is limited to the cases where each image only contains a single foreground object.\par
Perhaps the most relevant and competitive approaches to ours are the recent works of IIC~\cite{IIC} and AC~\cite{ARL}, which similarly uses the mutual information maximization concept to tackle unsupervised segmentation task. Different from these frameworks, our proposed method, inspired by~\cite{IMSAT}, models the unsupervised segmentation task as maximizing mutual information between the inputs and a discrete representation, \ie the cluster assignments.   
Moreover, while these methods rely on heuristic data augmentation strategies for their unsupervised model, our proposed framework incorporates a wisely customized adversarial training strategy as a learning-based data augmentation, which is jointly optimized with mutual information maximization objectives. Additionally, we suggest a new overlook into the problem by suggesting to directly cluster spatially connected groups of pixels, \ie superpixels, instead of the pixels. This strategy facilitates training, reduces the computational cost, and increases statistics' reliability during training by increasing batch size.

\section{Proposed method} \label{sec:PrMe}
\subsection{Overview}
\revised{The} semantic segmentation task can be considered as a mapping function $F:X\rightarrow Y$ from input image $X\in \mathbb{R}^{(w\times h\times c)}$ to the output map $Y\in \mathbb{B}^{(w\times h\times k)}$ where $w$ and $h$ define the input image's width and height and $c$ and $k$ represent the color and the semantic label channels of the input and the output, respectively. Note that in this representation, $Y$ is a discrete binary tensor, and the label of each pixel in the image $X$ is encoded as a one-hot binary vector in $Y$. Therefore, a group of pixels with an identical one-hot binary code represents the same semantic object class. Without loss of generality, we can assume there always exists a proper mapping function between the inputs and outputs, which can be estimated using a deep neural network\footnote{ This claim can be supported by this fact that deep learning has favorably applied to the supervised semantic segmentation task, and many algorithms successfully tackled this problem \cite{DeepLab, RefineNet, PSPNet, FCN}}. \par

\begin{figure}[t]
\begin{center}
\includegraphics[width=1\linewidth]{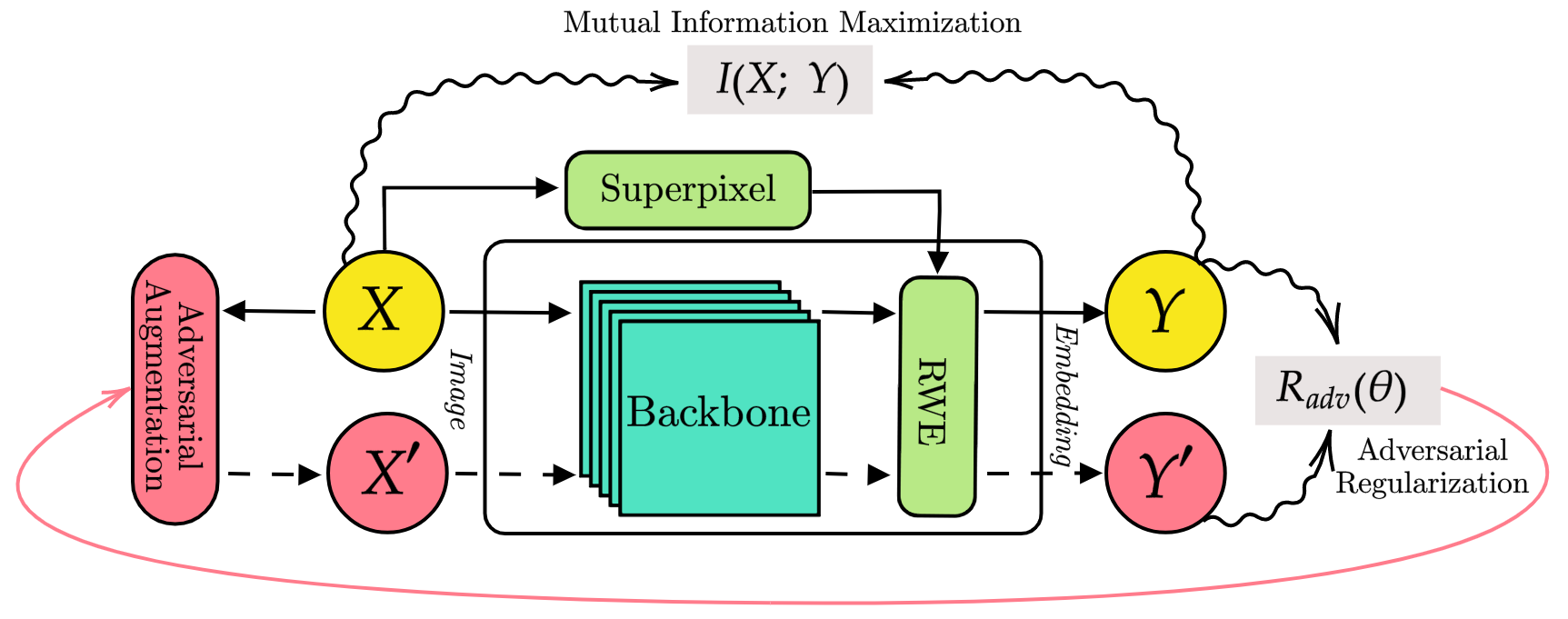}
\vspace{-2.5em}
\end{center}
   \caption{Building blocks of our proposed method. The \colorbox{myYellow}{yellow} and \colorbox{myRed}{red} colors indicate original data and augmented version of original data, respectively. The input image $X$ and its adversarial augmented version $X'$ are passed from the network and generate response $Y$ and $Y'$, respectively. The network is combined of three modules, Backbone, Superpixel, and RWE. Backbone is a fully convolutional network. Superpixel module extract sub-region from the input image. RWE map output of Backbone to the sub-regions to generate fixed-length feature. The clustering loss $I(X;Y)$ is computed between input $X$ and its corresponding embedding $Y$. The regularization loss $R(\theta)$ controls the network to generate rich embedding via \textit{Adversarial Augmentation} module.}
\label{fig:usn}
\end{figure}

In an unsupervised setting, the main task is to find a proper function $F$ using a deep model, which learns to perform this mapping without any annotation. To tackle this, we build a framework (\Fig~\ref{fig:usn}) based on three major components, including Region-Wise Embedding (RWE), Mutual Information Maximization (MIM), and Adversarial Regularization, which are separately elaborated in the followings.\par
\textbf{From Pixel to Region}: 
At the pixel-level, extracting semantically meaningful features is more complicated than the image-level, as we need to ensure maintaining the spatial smoothness and connectivity between the semantic label of neighboring pixels while generating an embedding for each pixel.
To this end, we propose to exploit segmentation region proposals, \eg superpixels, and to assign the same label to all the pixels in each region to ensure holding their spatial label smoothness while reducing the problem's complexity. These superpixels can be directly extracted using any unsupervised, model-based, or deep learning-based frameworks~\cite{SLIC, SSN, SP_Affinity_Loss, SP-FCN}. In fact, each superpixel provides a hypothetically good initial guess for the spatial connectivity of the semantically meaningful pixels in proximity without any supervision. Therefore, from each of these superpixel regions, one feature embedding can be extracted as the representative. The module extracting these features is called Region-Wise-Embedding (RWE) in our framework. RWE receives an embedding map from the backbone's output and produces a fixed-length embedding for each superpixel using a pooling strategy. The effect of the superpixel parameter selection and pooling strategy for this module will be discussed in the ablation study. Note that the RWE module might resemble the Region Proposal Network (RPN) in Faster R-CNN detector~\cite{FasterRCNN}.\par

\textbf{Mutual Information Maximization and Adversarial Regularization}: Training a deep network requires optimizing a loss function. In unsupervised settings, a loss function should optimize the network parameters without any annotated data. Hence, the design of the loss function capturing a proper source of information for the task plays a remarkable role in tackling the problem. In the absence of an annotated data, the input measurements and the model predictions are the only available source of information for training the model. In addition, data augmentation, as a very popular strategy, can be conducive to the unsupervised training scheme, \eg by generating additional pseudo labeled data. This is because the augmented version of input can hypothetically have the same label as the original input. Using this pair match, the model is encouraged to generate a similar output for this pair, \ie the input data, and its augmented version.\par
Selecting a proper data augmentation strategy with a reasonable perturbation/augmentation magnitude has a crucial impact on the quality of optimization convergence, \eg simple data augmentation or a minor perturbation may not help the optimization to converge to a good solution, and a high perturbation or incorrect augmentation strategy may result in optimization divergence,~\cite{dataAug}. To this end, the previous works~\cite{IIC, UDA} have also attempted to adopt a proper augmentation strategy for their methods to generate reasonable results. Despite this popularity, designing proper augmentation is still challenging since the perturbation should not alter the meaning of data.\par
To address this challenge, we utilize adversarial training in our data augmentation module in order to generate adversarial examples while bound the amount of perturbation in a safe region~\cite{IMSAT}. A safe region is an area where adding perturbation does not alter the meaning of input data, and it is a hyper-parameter in adversarial training. As depicted in Figure.~\ref{fig:usn}, the adversarial augmentation module gets an image $X$ and generate its augmented version $X^{'}$. This secondary input passes through the network and result in $Y^{'}$. The adversarial objective is then computed as a deviation between $Y^{'}$ and $Y$. Mutual information is an information-theoretic criterion to evaluate the correlation between two variables, so it is a valuable measure to train a deep network in the lack of annotation. We exploit this concept to maximize the dependency between input and output and encourage the model to hold maximum information between them. We show this criterion in the Figure \ref{fig:usn} between $X$ and $Y$.
The final loss function is a weighted average of the adversarial and mutual information terms.

\subsection{Method} This paper proposes a novel approach to find the proper mapping function called \textbf{In}formation \textbf{M}aximization and \textbf{A}dversarial \textbf{R}egularization \textbf{S}egmentation (\textbf{InMARS}) to segment input image into a predefined number of clusters without any label or supervision. To explain the pipeline, we start by providing further details about RWE and the variant of the extracted features. Then, as our clustering technique, we elaborate more on the Mutual Information Maximization objective. Finally, we present our customized adversarial training as a regularizer .

\textbf{Region Wise Embedding (RWE)}:
As mentioned earlier, in this stage, our goal is to produce candidate region/sub-regions to preserve spatial information to some extent and to generate smooth label predictions for each sub-region while generating a reliable feature from this candidate for the next stage. Moreover, we need to ensure that this RWE module can extract a fixed-length feature representation from each superpixel with arbitrary sizes before inputting them into the clustering objective. To this end, we evaluate \revised{three} different strategies to generate the fixed-length embedding from variable-size inputs as follow: (1) Sub-region Pooling, (2) Spatial Pyramid Pooling (SPP)~\cite{SPP}, and (3) Interpolation.\\ \revised{In the sub-region pooling strategy, we produce a single embedding by applying max/average-pooling to each superpixel separately. SPP is a fixed-length map that tunes 2D max-pooling parameters based on input size in multiple scales to generate a fixed size output for any input size's region. We utilize a zero mask in the SPP to zero out pixels' embedding that fall into SPP's rectangular region but do not belong to the selected superpixel. Then, we passed SPP's output from a series of fully connected (FC) layers to generate the final embedding. Interpolation is another strategy to build an equal size embedding for all candidates. For this, we resize superpixel regions into a predefined size. Then, similar to SPP, we apply a zero mask and feed a flattened version of these equal-sized regions to the fully connected layers.} Their effect on the performance will be studied in the ablation part.

\textbf{Unsupervised Segmentation by Maximizing Mutual Information and Adversarial Training}: 
Let $Y_i\in \mathbb{B}^{(w\times h)}$ be $i^{th}$ channel of $Y$ as a discrete representation in the network's output where $1\leq i \leq k$ and $k$ is the maximum number of the semantic labels/clusters. Also, let $x \sim X$ and $y_i \sim Y_i$ be random variables for the input and output, respectively. Given $x$ and considering each label as a conditionally independent variable, we can formulate the joint distribution of the output by $p_{\theta}\left(y_{1},\ldots, y_{k} \mid x\right)=\prod_{i=1}^{k} p_{\theta}\left(y_{i} \mid x\right)$, where $\theta$ is the parameters of the model, \ie deep neural network. During training, mutual information between input $X$ and output $Y$ is maximized using marginal entropy and conditional entropy, while the training is regularized via adversarial training. To this end, we have the following objective to minimize:

\vspace*{-0.75em}
\begin{equation} \label{eq:mainLoss}
    \revised{R(\theta)}-\lambda I(X;Y)
\end{equation}
\vspace*{-1.0em}

Where \revised{$R(\theta)$} is the regularization term that enforces the deep network to produce informative and reliable features. $I(X; Y )$ is the mutual information between input and embedding of a network, and $\lambda$ is a trade-off between two terms.

\underline{\emph{Mutual Information Objective}}: 
There are two ways to deal with the data clustering problem using mutual information maximization (MIM), 1) $MIM(X, Y)$ between the inputs $X$ and the outputs $Y$ and 2) $MIM(Y,Y^{\prime})$ between the output $Y$ and the prediction of an augmented version of the input $\hat{Y}$. In both approaches, the goal is to train a network that maps similar inputs into similar representations~\cite{IMSAT,IIC}. The $MIM(Y,Y^{\prime})$ requires a pair of the original input and augmented version, yet the $MIM(X, Y)$ does not depend on any additional pairs and can be readily computed from the network's output. Following~\cite{IMSAT}, we selected $MIM(X, Y)$ strategy as our strategy. In terms of entropy, mutual information can be described as $I(X; Y)=H(Y)-H(Y\mid X)$, where $H(Y)$ and $H(Y\mid X)$ are entropy and conditional entropy, respectively, defined by:

\vspace*{-0.75em}
\begin{equation} \label{eq:8}
H(Y) \equiv h(p_{\theta}(y)) = h\left(\frac{1}{N} \sum_{j=1}^{N} p_{\theta}\left(y \mid x_{j}\right)\right)
\end{equation}
\vspace*{-0.75em}
\begin{equation} \label{eq:9}
H(Y \mid X)=\frac{1}{N} \sum_{j=1}^{N} h\big(p_{\theta}\left(y \mid x_{j}\right)\big)
\end{equation}
\vspace*{-1.0em}

where $h(.)$ is the entropy function and $x_j$ is each training data input. \revised{According to \cite{Gomes_RIM}, we can incorporate prior knowledge, \eg a distribution over the sample size in each class in the case of a class imbalance in the data, by rewriting $H(Y)$ as $log(K) - KL[p_{\theta}(y)|| q(y)]$ where $K$ is the number of clusters, $q(y)$ is the prior distribution and $KL[.||.]$ is the Kullback-Leibler divergence~\cite{KLD}. In our experiments for simplicity, we assume the prior $q(y)$ is a uniform distribution.} \par
To better understand $MIM$ as an objective function, we could evaluate marginal entropy $H(Y)$ and conditional entropy $H(Y\mid X)$. The marginal entropy pushes cluster assignment to be uniform (distributing data to all clusters) while the conditional entropy forces cluster assignment to be explicit (push the same data to a single cluster).\par

\underline{\emph{Adversarial Training}}: To augment each data sample with adversarial training,  an additional regularization term $R(\theta)$ is incorporated into the loss function:

\vspace*{-0.75em}
\begin{equation} \label{eq:adv}
\revised{R(\theta ;\ T)}= KL\left(p_{\theta}\left(y_{i} \mid x\right), p_{\theta}\left(y_{i} \mid T(x)\right)\right)
\end{equation}
\vspace*{-1.0em}

where $KL,~p_{\theta}\left(y_{i} \mid x\right)$ and $p_{\theta}\left(y_{i} \mid T(x)\right)$ are KL divergence, the network's embedding for the original data $x$, and the transformed (augmented) data $T(x)$, respectively.\par
This regularization objective penalizes embedding dissimilarity between data and augmented versions and enforces the consistency of embedding on the network's output. We select pixel noise to increase the invariance of the extracted features against perturbations \revised{$r$ as $T(x) = x + r$.} Then, we regularize the network against adversarial samples while keeping the decision boundary in the low-density region of data distribution by confining adversarial perturbations. The confined adversarial samples are mined according to the following criteria.

\vspace*{-0.75em}
\begin{equation} \label{eq:t}
r=\underset{r^{'}}{\operatorname{argmax}}\left\{R_{a d v}(\theta ;\ x, T(x)) ; \|r^{'}\|_{2} \leq \varepsilon\right\}
\end{equation}
\vspace*{-1.0em}

Where $r$ is the additive noise \revised{and $\varepsilon$ is a hyper-parameter that controls the range of the local perturbation}. We can approximate adversarial term $r$ within the training process by an additional forward/backward pass. \revised{To this end, we called $R_{a d v}$ as adversarial form of $R$ when we compute $T(x)$ in adversarial fashion.} In addition, we also \revised{add} random affine perturbations to include geometrical invariance to the network as in \cite{IMSAT}.

\underline{\emph{Training Objective}}: Combining adversarial and mutual information terms result in the final objective function. Recall that our main objective (Equation \ref{eq:mainLoss}) consist of two terms $R_{a d v}(\theta)$ and $I(X;Y)$. Therefore, we can write ${\mathcal{L}}_{R}$ and ${\mathcal{L}}_{MI}$ for these terms as follow:
\vspace*{-0.5em}
\begin{equation} \label{eq:loss}
\begin{split}
    \revised{
            \mathcal{L}_{R}} & \revised{= \frac{1}{2} R_{a d v}+\frac{1}{2} R_{g e o}
            } \\
    {\mathcal{L}}_{MI} & = H(Y\mid X) - \gamma H(Y)\\
    {\mathcal{L}}_{total} & = \revised{\mathcal{L}_{R}} + \lambda {\mathcal{L}}_{MI}
\end{split}
\end{equation}

where \revised{$\mathcal{L}_{R}$} and ${\mathcal{L}}_{MI}$ are \revised{regularization} loss and mutual information loss, respectively. \revised{$\lambda$ and $\gamma$ are the hyper-parameters. We fix the $\lambda$ during training but decrease the $\gamma$ gradually (see Sec. IV - Hyper-parameters, for more details). The higher value of $\gamma$ at initial stages helps to uniformly generate class in the output while the decay of this coefficient gradually assigns more importance on the explicit class generation in the final stages.} $R_{a d v}$ and \revised{$R_{g e o}$} are adversarial additive pixel noise and \revised{random affine perturbation, respectively. For $R_{g e o}$ we use random scaling, translation, rotation and shearing similar to \cite{IMSAT}.} \revised{All hyper-parameters are tuned in a non-overlapping subset with the train and validation sets}, and the computed parameters are used for the final training and evaluation. Minimizing loss function ${\mathcal{L}_{total}}$ is equivalent to minimizing \revised{$R$} and maximizing $I(X;Y)$, simultaneously.\par

\section{Experiments} \label{sec:Ex}
\textbf{Dataset}: Ji \etal~\cite{IIC} has released two benchmark from COCO-Stuff~\cite{Cocostuff} and Potsdam~\cite{Potsdam} datasets, which has become the most popular datasets for evaluating the existing unsupervised image segmentation methods. Potsdam dataset consists of 8550 RGBIR $200\times200$ pixel satellite images with 3150 unlabelled samples. The first subset (Potsdam) has six classes (roads and cars, vegetation and trees, buildings, and clutter), while the second one (Potsdam-3) has three classes. Potsdam-3 is created by merging six classes of the first subset into three \revised{coarser} classes. COCO-Stuff~\cite{Cocostuff} is another challenging dataset that contains stuff classes. The first subset of COCO-Stuff has 15-\revised{coarse} classes with 52K images. These images at least contain $75\%$ of stuff pixels. The second subset (COCO-Stuff-3) is generated by taking 36K images from the first subset with the sky, ground, and plant classes.\par

\textbf{Implementation Details}: We conducted all experiments with randomly initialized networks. We used Hourglass~\cite{hourglass} as the main backbone of our segmentation network; however, this backbone can be replaced by any other fully convolutional networks, \eg the UNet~\cite{UNet}, as shown in our ablation study (Table~\ref{AS-backbone:long}). Hourglass or “stacked hourglass” network is built on the successive steps of pooling and upsampling done to produce a final set of predictions. Each hourglass module is closely connected to fully convolutional networks, and the connection of modules could process spatial information at multiple scales for dense prediction~\cite{hourglass}. We used two stacks of the hourglass with two residual blocks in each hourglass as our backbone. Then, the RWE head is employed on the top of the backbone to map pixels' embedding to regions. In RWE, we picked SLIC Superpixel \cite{SLIC} with parameters so that each candidate region conveys meaningful spatial information and has self-determining features. In fact, the form and spatial size of the ground truth classes can be important to choose proper superpixel parameters. Therefore, the parameters are set such that \emph{(i)} the number of superpixels is not less than the number of semantic classes; \emph{(ii)} Since under a general superpixel parameters' setup, there is no guarantee that each generated superpixel region carries the pixels with the same semantic labels, over-segmentation is essential in our framework. By over-segmentation, our framework can merge the superpixels with the same semantic classes. We implemented the RWE module with three different approaches:~(1) sub-region pooling, (2) SPP, and (3) interpolation. In the ablation part, we studied the impact of each approach on the network's performance (Table~\ref{AS-RWE:long}). In our final result, we chose SPP as the best module for the RWE module.

Our proposed method (InMARS) has three variants, namely InMARS$^{b}$, InMARS and InMARS$^{+}$. InMARS$^{b}$ is the argmaxed outputs from the backbone component of our model during inference (\Fig~\ref{fig:usn}). Note that in our implementations, we ensured that the number of the backbone's output channels is equal to the number of ground-truth classes. InMARS is the inference output of our proposed model (\Fig~\ref{fig:usn}) including RWE module, trained with $2*N_{gt}$ superpixels. InMARS$^{+}$ is another variant (multi-scale) of InMARS, incorporating all three superpixel resolutions (Table~\ref{AS-SP:long}) during inference to generate multiple outputs for each pixel. Then, the final prediction is formed by pixel-wise averaging of the three embeddings. We compared the performance of our method against \cite{ IIC, ARL,Pedregosa2011scikit-learn, Lowe:2004:DIF:993451.996342, Doersch, isola2015learning, DeepCluster}. We also reported the results for a supervised baseline, trained using the UNet~\cite{UNet}, as the best performing backbone from the possible backbones we used in our experiment.\par

\begin{figure*}[htb]
    \centering
    \begin{subfigure}[t]{0.125\textwidth}
      \includegraphics[width=1\textwidth]{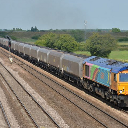}
    \end{subfigure}
    \begin{subfigure}[t]{0.125\textwidth}
      \includegraphics[width=1\textwidth]{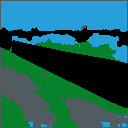}
    \end{subfigure}
    \begin{subfigure}[t]{0.125\textwidth}
      \includegraphics[width=1\textwidth]{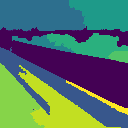}
    \end{subfigure}
    \begin{subfigure}[t]{0.125\textwidth}
      \includegraphics[width=1\textwidth]{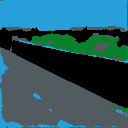}
     \end{subfigure}
    \begin{subfigure}[t]{0.125\textwidth}
      \includegraphics[width=1\textwidth]{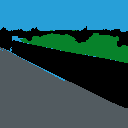}
    \end{subfigure}
    \begin{subfigure}[t]{0.125\textwidth}
      \includegraphics[width=1\textwidth]{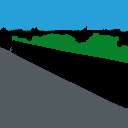}
    \end{subfigure} 
    \begin{subfigure}[t]{0.125\textwidth}
      \includegraphics[width=1\textwidth]{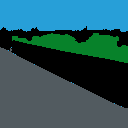}
    \end{subfigure}
    
    \par\smallskip
    \centering
    \begin{subfigure}[t]{0.125\textwidth}
      \includegraphics[width=1\textwidth]{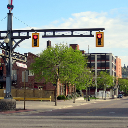}
    \end{subfigure}
    \begin{subfigure}[t]{0.125\textwidth}
      \includegraphics[width=1\textwidth]{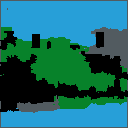}
    \end{subfigure}    
    \begin{subfigure}[t]{0.125\textwidth}
      \includegraphics[width=1\textwidth]{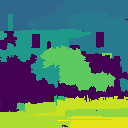}
    \end{subfigure}    
    \begin{subfigure}[t]{0.125\textwidth}
      \includegraphics[width=1\textwidth]{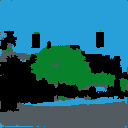}
     \end{subfigure}
    \begin{subfigure}[t]{0.125\textwidth}
      \includegraphics[width=1\textwidth]{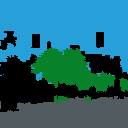}
    \end{subfigure}
    \begin{subfigure}[t]{0.125\textwidth}
      \includegraphics[width=1\textwidth]{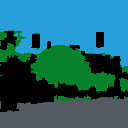}
    \end{subfigure}
    \begin{subfigure}[t]{0.125\textwidth}
      \includegraphics[width=1\textwidth]{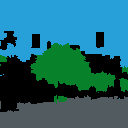}
    \end{subfigure}

    \par\smallskip
    \centering
    \begin{subfigure}[t]{0.125\textwidth}
      \includegraphics[width=1\textwidth]{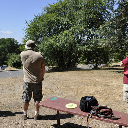}
    \end{subfigure}
    \begin{subfigure}[t]{0.125\textwidth}
      \includegraphics[width=1\textwidth]{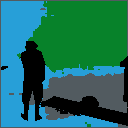}
    \end{subfigure}
    \begin{subfigure}[t]{0.125\textwidth}
      \includegraphics[width=1\textwidth]{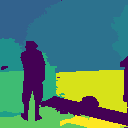}
    \end{subfigure}
    \begin{subfigure}[t]{0.125\textwidth}
      \includegraphics[width=1\textwidth]{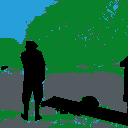}
     \end{subfigure}
    \begin{subfigure}[t]{0.125\textwidth}
      \includegraphics[width=1\textwidth]{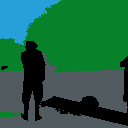}
    \end{subfigure}
    \begin{subfigure}[t]{0.125\textwidth}
      \includegraphics[width=1\textwidth]{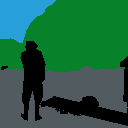}
    \end{subfigure}
    \begin{subfigure}[t]{0.125\textwidth}
      \includegraphics[width=1\textwidth]{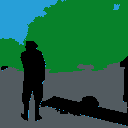}
    \end{subfigure}   
    
    \par\smallskip
    \centering
    \begin{subfigure}[t]{0.125\textwidth}
      \includegraphics[width=1\textwidth]{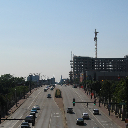}
    \end{subfigure}
    \begin{subfigure}[t]{0.125\textwidth}
      \includegraphics[width=1\textwidth]{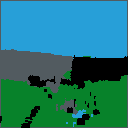}
    \end{subfigure}
    \begin{subfigure}[t]{0.125\textwidth}
      \includegraphics[width=1\textwidth]{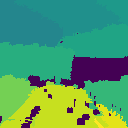}
    \end{subfigure}
    \begin{subfigure}[t]{0.125\textwidth}
      \includegraphics[width=1\textwidth]{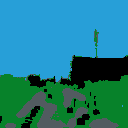}
     \end{subfigure}
    \begin{subfigure}[t]{0.125\textwidth}
      \includegraphics[width=1\textwidth]{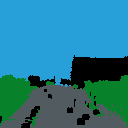}
    \end{subfigure}
    \begin{subfigure}[t]{0.125\textwidth}
      \includegraphics[width=1\textwidth]{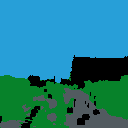}
    \end{subfigure}
    \begin{subfigure}[t]{0.125\textwidth}
      \includegraphics[width=1\textwidth]{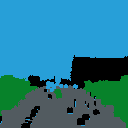}
    \end{subfigure} 
    
    \par\smallskip
    \centering
    \begin{subfigure}[t]{0.125\textwidth}
      \includegraphics[width=1\textwidth]{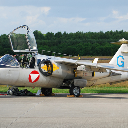}
      \subcaption{Input}
    \end{subfigure}
    \begin{subfigure}[t]{0.125\textwidth}
      \includegraphics[width=1\textwidth]{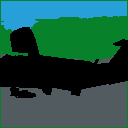}
      \subcaption{IIC\cite{IIC}}
    \end{subfigure}
    \begin{subfigure}[t]{0.125\textwidth}
      \includegraphics[width=1\textwidth]{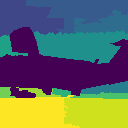}
      \subcaption{Superpixel}
    \end{subfigure}
    \begin{subfigure}[t]{0.125\textwidth}
      \includegraphics[width=1\textwidth]{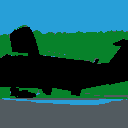}
      \subcaption{InMARS$^{b}$}
     \end{subfigure}
    \begin{subfigure}[t]{0.125\textwidth}
      \includegraphics[width=1\textwidth]{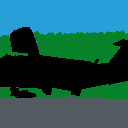}
      \subcaption{InMARS$^{+}$}
    \end{subfigure}
    \begin{subfigure}[t]{0.125\textwidth}
      \includegraphics[width=1\textwidth]{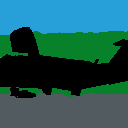}
      \subcaption{Supervised}
    \end{subfigure}    
    \begin{subfigure}[t]{0.125\textwidth}
      \includegraphics[width=1\textwidth]{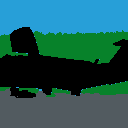}
      \subcaption{GT}
    \end{subfigure}
    
    \caption{\textbf{Segmentation Results from COCO-Stuff-3 \cite{Cocostuff}}. (a) Few input images examples, (b) The IIC~\cite{IIC} outputs. The figures (c)-(e) represent the component of our model, \ie (c) Superpixels in the finest resolution (\ie $N_{gt}^{2}$), (d) The argmaxed version of the our backbone outputs ( InMARS$^{b}$) (e) The outputs of the multi-scale variant of our model (InMARS$^{+}$). The outputs of the supervised baseline and the Ground-truths are shown in (f) and (g), respectively.}
    \vspace{-1em}
    \label{fig:vis:coco-st3}
  \end{figure*}

\textbf{Hyper-parameters}: All the hyper-parameters in our experiments were tuned in a non-overlapping subset of the dataset (\%10 randomly selected), and they were fixed during optimization and evaluation on the dataset on which we evaluated and reported the final results. We selected Adam optimizer~\cite{Adam} with the initial learning rate $lr_{init}=10^{-5}$ and exponential decays for the learning rate by $\alpha=0.96$ in every epoch to do all the experiments. The learning rate in every epoch is derived by $lr_{i} = \alpha^{i}*lr_{init}$. In this equation $lr_{init}$ is the initial learning rate, $i$ is the numerator of epoch and $lr_{i}$ is the learning rate in the epoch $i$. We trained the network for 100 epochs and picked the last model for evaluation. We selected ($5$, $2*N_{gt}$) for compactness and number of SLIC superpixels during training. To speed up the training process, we extracted all superpixels once offline. In training, invalid pixels (pixels that do not belong to any of the classes) were discarded, and the loss function was calculated just on valid pixels. We followed the prior works for image size ($200\times200$ and $128\times128$ pixel images for Potsdam and COCO-Stuff, respectively. For hourglass, we resized the network's inputs from $200\times200$ to $192\times192$ pixels, as with depth equal to 4, the input should be divisible by 16 ($2^{depth}$). We used PyTorch library~\cite{PyTorch} for implementation and NVidia Titan X and GTX 1080ti GPUs for training. The hyper-parameters $\lambda$, $\gamma$ and $\varepsilon$ are set to $4$, $1$ and $.25$, respectively. We fixed the $\lambda$ and $\varepsilon$ during training but decreased the $\gamma$ from $1$ to $0.1$.\par

\subsection{Evaluation}
As in IIC~\cite{IIC} and AC~\cite{ARL}, we used the same evaluation measure (pixel accuracy) to report the performance of our method, \ie $\label{eq:ACC}
ACC=\max _{m} \frac{\sum_{n=1}^{N} 1\left\{l_{n}=m\left(c_{n}\right)\right\}}{N}$
where $l_n$ and $c_n$ are the ground-truth label and cluster assignment for input $x_n$, respectively. $m$ ranges over all possible one-to-one mapping between cluster assignments and ground-truth classes. The maximum operator is effectively computed using linear assignment~\cite{Kuhn1955Hungarian} to find the best one-to-one mapping between the predicted classes and ground-truth labels. In both the assignment and evaluation step, we discarded invalid pixels as in visualization (Figure \ref{fig:vis:coco-st3}).

Table \ref{table-acc:long}. demonstrates the superior performance of our InMARS$^{+}$ compared to the state-of-the-art methods in the three datasets, validating the efficacy of our suggested modules for this task, \ie unsupervised segmentation.

\begin{table}
\begin{center}
\scriptsize

\caption{\textbf{Accuracy of Unsupervised Image Segmentation}}
\label{table-acc:long}
\label{table-acc:onecol}
\begin{tabular}{|l|c|c|c|c|}
\hline
& COCO & COCO & Potsdam-3 & Potsdam\\
& Stuff-3 & Stuff & & \\
\hline\hline
\revised{Supervised Baseline} & \revised{84.7} & \revised{89.7} & \revised{80.2} & \revised{82.4} \\
\hdashline
K-means \cite{Pedregosa2011scikit-learn}
& 52.2 & 14.1 & 45.7 & 35.3 \\
SIFT \cite{Lowe:2004:DIF:993451.996342}
& 38.1 & 20.2 & 38.2 & 28.5 \\
Doersch 2015 \cite{Doersch}
& 47.5 & 23.1 & 49.6 & 37.2 \\
Isola 2016 \cite{isola2015learning}
& 54.0 & 24.3 & 63.9 & 44.9 \\
Deep Cluster 2018 \cite{DeepCluster}
& 41.6 & 19.9 & 41.7 & 29.2 \\
IIC 2019 \cite{IIC}
& 72.3 & 27.7 & 65.1 & 45.4 \\
AC 2020 \cite{ARL}
& \underline{72.9} & \underline{30.8} & 66.5 & \textbf{49.3} \\
\hline
\textbf{InMARS$^{b}$} \textit{(Ours)}
& 72.2 & 29.2 & 68.5 & 46.2 \\
\textbf{InMARS} \textit{(Ours)}
& 72.8 & 30.3 &  \underline{68.8} & 46.9 \\
\textbf{InMARS$^{+}$} \textit{(Ours)}
&\textbf{73.1} &\textbf{31.0} &\textbf{70.1} & \underline{47.3} \\
\hline

\end{tabular}
\footnotesize{\\\textsuperscript{b} InMARS backbone with $N_{gt}$ output classes.}
\footnotesize{\\\textsuperscript{+} InMARS with multi-scale test by different number of superpixels.}
\end{center}
\end{table}
\par

\begin{figure}[!tb]
    \centering
    \begin{subfigure}[t]{0.11\textwidth}
      \includegraphics[width=\textwidth]{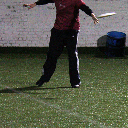}
    \end{subfigure}
    \begin{subfigure}[t]{0.11\textwidth}
      \includegraphics[width=\textwidth]{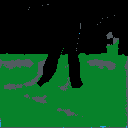}
     \end{subfigure}
    \begin{subfigure}[t]{0.11\textwidth}
      \includegraphics[width=\textwidth]{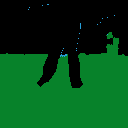}
    \end{subfigure}
    \begin{subfigure}[t]{0.11\textwidth}
      \includegraphics[width=\textwidth]{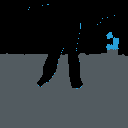}
    \end{subfigure}
    
    \par\smallskip
    \centering
    \begin{subfigure}[t]{0.11\textwidth}
      \includegraphics[width=\textwidth]{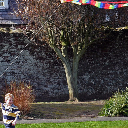}
      \subcaption{Input}
    \end{subfigure}
    \begin{subfigure}[t]{0.11\textwidth}
      \includegraphics[width=\textwidth]{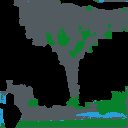}
      \subcaption{InMARS$^{b}$}
     \end{subfigure}
    \begin{subfigure}[t]{0.11\textwidth}
      \includegraphics[width=\textwidth]{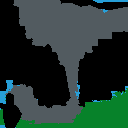}
      \subcaption{InMARS$^{+}$}      
    \end{subfigure}
    \begin{subfigure}[t]{0.11\textwidth}
      \includegraphics[width=\textwidth]{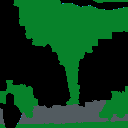}
      \subcaption{GT}
    \end{subfigure}
    \caption{\textbf{Failure Cases}. Some failure cases of the proposed method from COCO-Stuff-3 \cite{Cocostuff}. The proposed algorithm fails to recognize some classes due to the similarity of classes (first row) and clutter in the background (second row)}
    \label{fig:failure:coco-st3}
\end{figure}

\textbf{Qualitative Result}: Figure \ref{fig:vis:coco-st3}. illustrates the output of InMARS (output of the backbone) and InMARS$^{+}$ (output of the RWE) on the COCO-Stuff-3 dataset. The RWE module has spatial smoothness ability and performs well on the top of the backbone to produce a continuous segmentation mask while preserving semantic information. The superpixels with the finest resolution ($N_{gt}^2$) are displayed in \Fig~3-(c). In addition, we included the output of the fully supervised UNet~\cite{UNet} and IIC~\cite{IIC} as two baselines for visual comparison. Note that similar to~\cite{IIC, UNet} in all the results, invalid pixels (pixels that do not belong to any of the classes) were discarded and are shown with dark colors. In Figure \ref{fig:failure:coco-st3} some failure cases presented. The first row is an ambiguous sample of the ground class (gray color in the ground-truth). Our method failed to distinguish this sample (artificial turf) from natural class. In fact, this level of perception requires a higher level of knowledge which is only available in the supervised objective. The second row shows a sample with a cluttered background. Our method failed to understand the natural class since it cannot understand this level of clutter.

\begin{figure}[!tb]
    
    \centering
    \crule[colorSky]{.15cm}{.15cm}
    \small{Sky}
    \crule[colorPlants]{.15cm}{.15cm}
    \small{Plants}
    \crule[colorGround]{.15cm}{.15cm}
    \small{Ground}
    
    \vspace*{0in}
    \begin{subfigure}[t]{0.115\textwidth}
    
     \includegraphics[width=\textwidth, trim={0.25cm 0.25cm 0 0},clip]{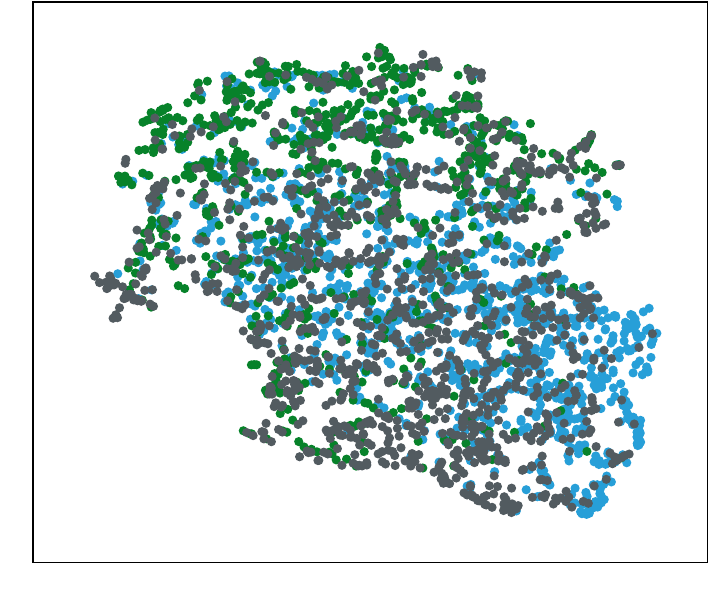}
    \vspace*{-\baselineskip}
    \end{subfigure}
    \begin{subfigure}[t]{0.115\textwidth}
      \includegraphics[width=\textwidth, trim={0.25cm 0.25cm 0 0},clip]{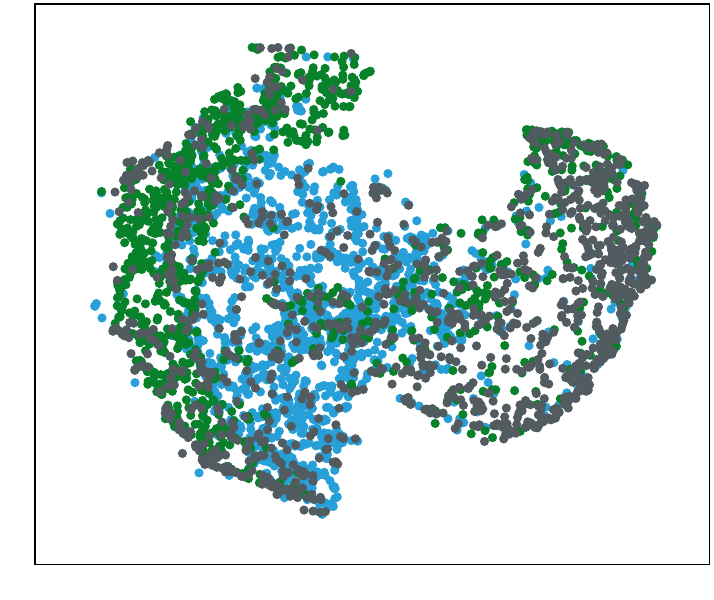}
    \vspace*{-\baselineskip}
    \end{subfigure}
    \begin{subfigure}[t]{0.115\textwidth}
      \includegraphics[width=\textwidth, trim={0.25cm 0.25cm 0 0},clip]{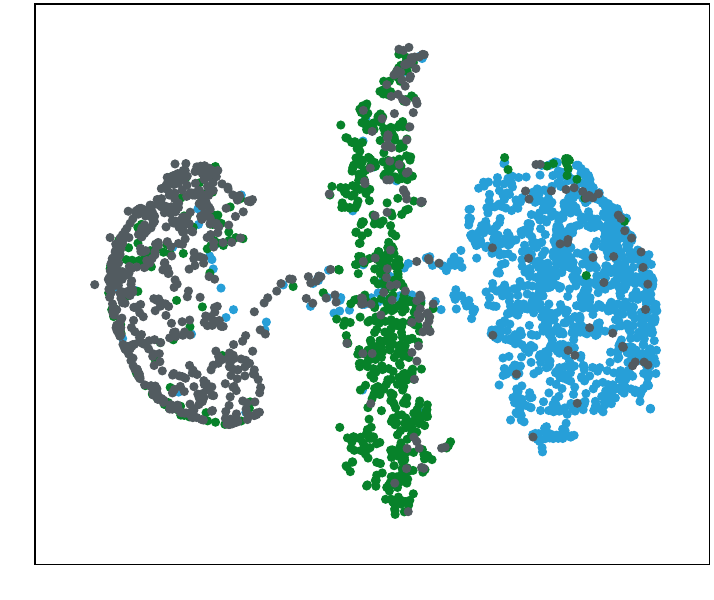}
    \vspace*{-\baselineskip}
    \end{subfigure}
    \begin{subfigure}[t]{0.115\textwidth}
      \includegraphics[width=\textwidth, trim={0.25cm 0.25cm 0 0},clip]{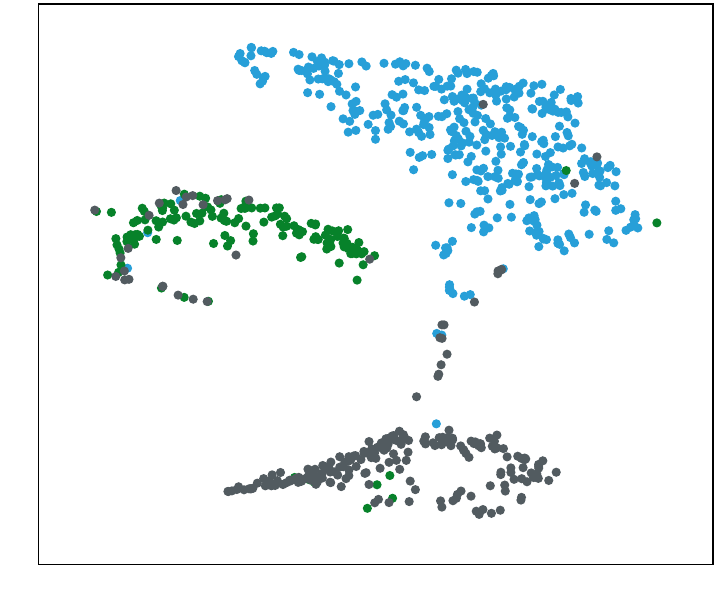}
    \vspace*{-\baselineskip}
    \end{subfigure}    
    \caption{\textbf{TSNE Visualization}. TSNE plot for the superpixel embedding on a subset of COCO-Stuff dataset using our proposed method (InMARS$^{+}$). From left to right, before training with random weights, after training, and trained model for images with $IoU > .25$ and $IoU > .5$, respectively.}
    \label{fig:tsne:coco-st3}
  \end{figure}

\textbf{Cluster Visualization}: Figure \ref{fig:tsne:coco-st3} illustrated the TSNE visualization for superpixels' embedding in the output of the network. It is worth to mention that TSNE is a 2D visualization that map high-dimensional data to lower dimension. This mapping cannot reflect the high dimensional separation capability of the method solely. However, we tried to highlight the network's performance by reducing the image with low average Intersection-over-Union (IoU) with ground-truth. Figure \ref{fig:tsne:coco-st3} (a) is the network's output at the initial stage, before training. As we expected, samples fall in a random pattern. Figure \ref{fig:tsne:coco-st3} (b) is the network's output after training. Figure \ref{fig:tsne:coco-st3} (c) and (d) are obtained by including image with IoU more than .25\% and .50\%, respectively. The TSNE of the trained model in Figure \ref{fig:tsne:coco-st3} (c) and (d) confirm that InMARS$^{+}$ can separate classes in feature space properly.

\subsection{Ablation Study}
In this part, we assessed the role of each component in the proposed method. Table \ref{AS-backbone:long} - Table \ref{AS-SP:long} present the ablation study of different components conducted on 10\% of randomly selected images of datasets.\par


\begin{table}[!tb]
\centering
\scriptsize
\vspace{-0.5em}
\captionof{table}{\textbf{Backbone}.We trained different backbone architectures with the same loss function and RWE parameters in the reduced version of COCO-Stuff-3 and Potsdam-3 datasets to compare their effects on the proposed method.}
\label{AS-backbone:long}
\label{AS-backbone:onecol}
\begin{tabular}{|l|c|c|}
\hline
& COCO-Stuff-3 & Potsdam-3\\
\hline\hline
UNet \cite{UNet} & 52.5 & 50.2 \\
Hourglass \cite{hourglass} & \textbf{54.5} & \textbf{51.4} \\
\hline
\end{tabular}
\end{table}

\begin{table}[!tb]

\centering
\scriptsize
\captionof{table}{\textbf{Region-Wise Embedding \textit{(RWE)}}. We tested three configurations for the RWE head on the top of the backbone to build a fixed-length embedding from superpixels. Max and average pooling are two kinds of sub-region pooling. SPP is spatial pyramid pooling, and FC stands for fully connected layers. The heads with the FC layer perform better since they can model complex tasks.}
\label{AS-RWE:long}
\label{AS-RWE:onecol}
\begin{tabular}{|l|c|c|}
\hline
& COCO-Stuff-3 & Potsdam-3\\
\hline\hline
max-pooling & 54.5 & 51.4 \\
mean-pooling & 50.1 & 51.0 \\
SPP~\cite{SPP}+FC & 57.3 & \textbf{55.5} \\
Interpolation+FC & \textbf{59.8} & 55.4 \\
\hline
\end{tabular}

\end{table}

\begin{table}[!tb]

\centering
\scriptsize
\revised{\captionof{table}{\textbf{Adversarial Regularization}. we illustrated the accuracy of the trained network \revised{without any regularization ($no-adversarial$)}, with \revised{adversarial} additive noise \revised{($R_{a d v}$)}, \revised{random geometrical} affine perturbation \revised{($R_{g e o}$)} and, their combination. The backbone is hourglass and \revised{max-pooling} module selected for RWE part.}
\label{AS-ADV:onecol}
\label{AS-ADV:long}
\begin{tabular}{|l|c|c|}
\hline
& COCO-Stuff-3 & Potsdam-3\\
\hline\hline
$no-adversarial$& 44.2 & 42.3 \\
\revised{$R_{a d v}$}& 54.5 & 51.4 \\
\revised{$R_{g e o}$}& 61.2 & 59.1 \\
\revised{$\frac{1}{2} R_{a d v}+\frac{1}{2} R_{g e o}$} & \textbf{61.6} & \textbf{60.4} \\
\hline
\end{tabular}}
\end{table}


\begin{table}[!tb]
\centering
\scriptsize
\captionof{table}{\textbf{Superpixel}. we inspected the effect of number of superpixel. $(N_{gt})$ is the number of ground-truth classes.}
\label{AS-SP:long}
\begin{tabular}{|l|c|c|}
\hline
(number of superpixels) & COCO-Stuff-3 & Potsdam-3\\
\hline\hline
$N_{gt}$ & 54.5 & 51.4 \\
$2*N_{gt}$ & \textbf{55.3} & 52.1 \\
$N_{gt}^{2}$ & 55.2 & \textbf{52.6} \\
\hline
\end{tabular}
\end{table}

\textbf{Backbone}: The backbone is a fully convolutional deep network, like FCN \cite{FCN} which gets the input image and generates an embedding map of equal size to the input image. In Table~\ref{AS-backbone:long}, we compared the effect of choosing different backbone architectures on the accuracy of the proposed method in the Potsdam-3 and COCO-Stuff-3 dataset. We chose UNet \cite{UNet} and Hourglass \cite{hourglass} backbone in our experiments, while other semantic segmentation backbones can be used instead. The results depicted that the hourglass backbone is performed best among others.\par

\textbf{RWE}: 
We designed three different ways to generate fixed-length embedding for superpixels: (1) Sub-region pooling, (2) SPP, and (3) interpolation. 
In Table~\ref{AS-RWE:long}, we reported the result of experiments with different RWE modules. In the InMARS, we chose SPP+FC for our final model.\par

\textbf{Adversarial Regularization}: The regularization term plays a remarkable role in the performance of the proposed method. Table~\ref{AS-ADV:long}. compares two \revised{different} strategies for regularization: \revised{adversarial} additive noise and \revised{random geometrical} affine perturbation (similar to~\cite{IMSAT}). The best accuracy achieved by a combination of two terms according to the Equation $\revised{\frac{1}{2} R_{a d v}+\frac{1}{2} R_{g e o}}$.
  
\textbf{Superpixel}: Table \ref{AS-SP:long}. shows various parameter selection for superpixel candidates. The extent and size of target classes define the approximate number of superpixels. The number of superpixels is selected regarding the number of ground-truth classes $(N_{gt})$. Since superpixel is not an ideal proposal, over-segmentation performs better. Finally, we selected combination of $N_{gt}$, $2N_{gt}$ and $N_{gt}^{2}$ in InMARS$^{+}$.\par

%
%


\section{Conclusion}
We have proposed a fully unsupervised semantic segmentation network, which utilizes the mutual information as the supervision and adversarial training as a regularization. The proposed method leads to an unsupervised semantic segmentation model that can partition the scene into semantic classes while generating good boundaries in the output by exploiting the off-the-shelf superpixel method. We further demonstrate the state-of-the-art results on benchmarks comparable with supervised methods and wish to reduce the gap with these methods. In the future, improvement of the method to handle more semantic classes in complex datasets would be fruitful.

\ifCLASSOPTIONcaptionsoff
  \newpage
\fi



%
\bibliographystyle{IEEEtran}
\bibliography{IEEEexample}

\begin{thebibliography}{10}
\providecommand{\url}[1]{#1}
\csname url@rmstyle\endcsname
\providecommand{\newblock}{\relax}
\providecommand{\bibinfo}[2]{#2}
\providecommand\BIBentrySTDinterwordspacing{\spaceskip=0pt\relax}
\providecommand\BIBentryALTinterwordstretchfactor{4}
\providecommand\BIBentryALTinterwordspacing{\spaceskip=\fontdimen2\font plus
\BIBentryALTinterwordstretchfactor\fontdimen3\font minus
  \fontdimen4\font\relax}
\providecommand\BIBforeignlanguage[2]{{%
\expandafter\ifx\csname l@#1\endcsname\relax
\typeout{** WARNING: IEEEtran.bst: No hyphenation pattern has been}%
\typeout{** loaded for the language `#1'. Using the pattern for}%
\typeout{** the default language instead.}%
\else
\language=\csname l@#1\endcsname
\fi
#2}}

\bibitem{DeepLab}
L.-C. Chen, G.~Papandreou, I.~Kokkinos, K.~Murphy, and A.~L. Yuille, ``Deeplab:
  Semantic image segmentation with deep convolutional nets, atrous convolution,
  and fully connected crfs.'' \emph{T-PAMI}, 2018.

\bibitem{RefineNet}
G.~Lin, A.~Milan, C.~Shen, and I.~Reid, ``Refine{N}et: {M}ulti-path refinement
  networks for high-resolution semantic segmentation,'' in \emph{CVPR}, 2017.

\bibitem{HRNet}
J.~Wang, \emph{et~al.}, ``Deep high-resolution representation learning for
  visual recognition,'' \emph{T-PAMI}, 2019.

\bibitem{PSPNet}
H.~{Zhao}, J.~{Shi}, X.~{Qi}, X.~{Wang}, and J.~{Jia}, ``Pyramid scene parsing
  network,'' in \emph{CVPR}, 2017.

\bibitem{FCN}
J.~Long, E.~Shelhamer, and T.~Darrell, ``Fully convolutional networks for
  semantic segmentation,'' in \emph{CVPR}, 2015.

\bibitem{ARL5}
M.~Caron, P.~Bojanowski, A.~Joulin, and M.~Douze, ``Deep clustering for
  unsupervised learning of visual features.'' in \emph{ECCV}, 2018.

\bibitem{ARL17}
R.~Girshick, J.~Donahue, T.~Darrell, and J.~Malik, ``Rich feature hierarchies
  for accurate object detection and semantic segmentation,'' in \emph{CVPR},
  2014.

\bibitem{IIC38}
T.-Y. Lin, M.~Maire, S.~Belongie, J.~Hays, P.~Perona, D.~Ramanan,
  P.~Doll{\'a}r, and C.~L. Zitnick, ``Microsoft coco: Common objects in
  context,'' in \emph{ECCV}, 2014.

\bibitem{DEC}
J.~Xie, R.~Girshick, and A.~Farhadi, ``Unsupervised deep embedding for
  clustering analysis,'' in \emph{ICML}, 2016.

\bibitem{VaDE}
Z.~Jiang, Y.~Zheng, H.~Tan, B.~Tang, and H.~Zhou, ``Variational deep embedding:
  An unsupervised and generative approach to clustering,'' in \emph{IJCAI},
  2017.

\bibitem{InfoGAN}
X.~Chen, Y.~Duan, R.~Houthooft, J.~Schulman, I.~Sutskever, and P.~Abbeel,
  ``Infogan: Interpretable representation learning by information maximizing
  generative adversarial nets,'' in \emph{NeurIPS}, 2016.

\bibitem{IMSAT}
W.~Hu, T.~Miyato, S.~Tokui, E.~Matsumoto, and M.~Sugiyama, ``Learning discrete
  representations via information maximizing self-augmented training,'' in
  \emph{ICML}, 2017.

\bibitem{AE+CE}
K.~G. Dizaji, A.~Herandi, C.~Deng, W.~Cai, and H.~Huang, ``Deep clustering via
  joint convolutional autoencoder embedding and relative entropy
  minimization,'' in \emph{ICCV}, 2017.

\bibitem{JULE}
J.~{Yang}, D.~{Parikh}, and D.~{Batra}, ``Joint unsupervised learning of deep
  representations and image clusters,'' in \emph{CVPR}, 2016.

\bibitem{DAC}
J.~Chang, L.~Wang, G.~Meng, S.~Xiang, and C.~Pan, ``Deep adaptive image
  clustering,'' in \emph{ICCV}, 2017.

\bibitem{Cocostuff}
H.~Caesar, J.~Uijlings, and V.~Ferrari, ``Coco-stuff: Thing and stuff classes
  in context,'' in \emph{CVPR}, 2018.

\bibitem{WNet}
X.~Xia and B.~Kulis, ``W-net: {A} deep model for fully unsupervised image
  segmentation,'' \emph{CoRR}, 2017.

\bibitem{IIC}
X.~Ji, J.~F. Henriques, and A.~Vedaldi, ``Invariant information clustering for
  unsupervised image classification and segmentation,'' in \emph{ICCV}, 2019.

\bibitem{SegSort}
J.-J. Hwang, S.~X. Yu, J.~Shi, M.~D. Collins, T.-J. Yang, X.~Zhang, and L.-C.
  Chen, ``Segsort: Segmentation by discriminative sorting of segments,'' in
  \emph{ICCV}, 2019.

\bibitem{ARL}
Y.~Ouali, C.~Hudelot, and M.~Tami, ``Autoregressive unsupervised image
  segmentation,'' in \emph{ECCV}, 2020.

\bibitem{Potsdam}
Potsdam, ``\uppercase{ISPRS}. \uppercase{ISPRS 2D} \uppercase{S}emantic
  \uppercase{L}abeling \uppercase{C}ontest.
  \url{http://www2.isprs.org/commissions/comm3/wg4/semantic-labeling.html}.''

\bibitem{ReviewSS2020}
S.~Minaee, Y.~Boykov, F.~Porikli, A.~J. Plaza, N.~Kehtarnavaz, and
  D.~Terzopoulos, ``Image segmentation using deep learning: A survey,''
  \emph{ArXiv:2001.05566}, 2020.

\bibitem{DANet}
J.~Fu, J.~Liu, H.~Tian, Y.~Li, Y.~Bao, Z.~Fang, and H.~Lu, ``Dual attention
  network for scene segmentation,'' in \emph{CVPR}, 2019.

\bibitem{He_2013_CVPR}
K.~He, F.~Wen, and J.~Sun, ``K-means hashing: An affinity-preserving
  quantization method for learning binary compact codes,'' in \emph{CVPR},
  2013.

\bibitem{GMM-Cl}
G.~J. McLachlan and K.~E. Basford, \emph{Mixture models: Inference and
  applications to clustering}.\hskip 1em plus 0.5em minus 0.4em\relax M. Dekker
  New York, 1988.

\bibitem{NIPS2004_2602}
L.~Xu, J.~Neufeld, B.~Larson, and D.~Schuurmans, ``Maximum margin clustering,''
  in \emph{NeurIPS}, 2004.

\bibitem{On-spectral-clustering}
A.~Y. Ng, M.~I. Jordan, and Y.~Weiss, ``On spectral clustering: Analysis and an
  algorithm,'' in \emph{NeurIPS}, 2001.

\bibitem{kmean+AE}
B.~Yang, X.~Fu, N.~D. Sidiropoulos, and M.~Hong, ``Towards k-means-friendly
  spaces: Simultaneous deep learning and clustering,'' in \emph{ICML}, 2017.

\bibitem{ARL4}
A.~Bielski and P.~Favaro, ``Emergence of object segmentation in perturbed
  generative models,'' in \emph{NeurIPS}, 2019.

\bibitem{ReDrawGAN}
M.~Chen, T.~Arti\`{e}res, and L.~Denoyer, ``Unsupervised object segmentation by
  redrawing,'' in \emph{NeurIPS}, 2019.

\bibitem{SLIC}
R.~{Achanta}, A.~{Shaji}, K.~{Smith}, A.~{Lucchi}, P.~{Fua}, and
  S.~{Süsstrunk}, ``Slic superpixels compared to state-of-the-art superpixel
  methods,'' \emph{T-PAMI}, 2012.

\bibitem{SSN}
V.~Jampani, D.~Sun, M.-Y. Liu, M.-H. Yang, and J.~Kautz, ``Superpixel sampling
  networks,'' in \emph{ECCV}, 2018.

\bibitem{SP_Affinity_Loss}
W.-C. Tu, M.-Y. Liu, V.~Jampani, D.~Sun, S.-Y. Chien, M.-H. Yang, and J.~Kautz,
  ``Learning superpixels with segmentation-aware affinity loss,'' in
  \emph{CVPR}, 2018.

\bibitem{SP-FCN}
F.~Yang, Q.~Sun, H.~Jin, and Z.~Zhou, ``Superpixel segmentation with fully
  convolutional networks,'' \emph{CVPR}, 2020.

\bibitem{FasterRCNN}
S.~Ren, K.~He, R.~Girshick, and J.~Sun, ``Faster r-cnn: Towards real-time
  object detection with region proposal networks,'' in \emph{NeurIPS}, 2015.

\bibitem{dataAug}
L.~Perez and J.~Wang, ``The effectiveness of data augmentation in image
  classification using deep learning,'' \emph{ArXiv:1712.04621}, 2017.

\bibitem{UDA}
Q.~Xie, Z.~Dai, E.~Hovy, T.~Luong, and Q.~Le, ``Unsupervised data augmentation
  for consistency training,'' in \emph{NeurIPS}, 2020.

\bibitem{SPP}
K.~He, X.~Zhang, S.~Ren, and J.~Sun, ``Spatial pyramid pooling in deep
  convolutional networks for visual recognition,'' in \emph{ECCV}, 2014.

\bibitem{Gomes_RIM}
A.~Krause, P.~Perona, and R.~Gomes, ``Discriminative clustering by regularized
  information maximization,'' in \emph{NeurIPS}, 2010.

\bibitem{KLD}
S.~Kullback and R.~A. Leibler, ``On information and sufficiency,'' \emph{Annals
  of Mathematical Statistics}, 1951.

\bibitem{hourglass}
A.~Newell, K.~Yang, and J.~Deng, ``Stacked hourglass networks for human pose
  estimation,'' in \emph{ECCV}, 2016.

\bibitem{UNet}
O.~Ronneberger, P.~Fischer, and T.~Brox, ``U-net: Convolutional networks for
  biomedical image segmentation,'' in \emph{MICCAI}, 2015.

\bibitem{Pedregosa2011scikit-learn}
F.~Pedregosa, \emph{et~al.}, ``Scikit-learn: Machine learning in python,''
  \emph{Journal of Machine Learning Research}, 2011.

\bibitem{Lowe:2004:DIF:993451.996342}
D.~G. Lowe, ``Distinctive image features from scale-invariant keypoints,''
  \emph{IJCV}, 2004.

\bibitem{Doersch}
C.~Doersch, A.~Gupta, and A.~A. Efros, ``Unsupervised visual representation
  learning by context prediction,'' in \emph{ICCV}, 2015.

\bibitem{isola2015learning}
P.~Isola, D.~Zoran, D.~Krishnan, and E.~Adelson, ``Learning visual groups from
  co-occurrences in space and time,'' \emph{ArXiv:1511.06811}, 2015.

\bibitem{DeepCluster}
M.~Caron, P.~Bojanowski, A.~Joulin, and M.~Douze, ``Deep clustering for
  unsupervised learning of visual features,'' in \emph{ECCV}, 2018.

\bibitem{Adam}
D.~P. Kingma and J.~Ba, ``Adam: A method for stochastic optimization,''
  \emph{CoRR}, vol. abs/1412.6980, 2015.

\bibitem{PyTorch}
A.~Paszke, S.~Gross, and G.~C. Soumith~Chintala, ``Pytorch: An open source
  machine learning framework. \url{https://pytorch.org/}.''

\bibitem{Kuhn1955Hungarian}
H.~W. Kuhn, ``{The Hungarian Method for the Assignment Problem},'' \emph{Naval
  Research Logistics Quarterly}, 1955.

\end{thebibliography}


%








\end{document}